\documentclass[conference]{IEEEtran}

\usepackage{graphicx}
\usepackage{amsmath}
\usepackage{float}
\usepackage{booktabs, makecell}
\usepackage{subcaption}
\usepackage{caption}
\usepackage{amsmath, amssymb, amsfonts}

\usepackage[T1]{fontenc}
\usepackage{cite}
\usepackage[top=0.75in, bottom=1.05in, left=0.73in, right=0.73in]{geometry}

\title{FlexFed: Mitigating Catastrophic Forgetting in Heterogeneous Federated Learning in Pervasive Computing Environments}

\author{
\IEEEauthorblockN{Sara Alosaime}
\IEEEauthorblockA{\textit{University of Warwick} \\
Sara.Alosaime@warwick.ac.uk}
\and
\IEEEauthorblockN{Arshad Jhumka}
\IEEEauthorblockA{\textit{University of Leeds} \\
H.A.Jhumka@leeds.ac.uk}
}

\begin{document}

\maketitle

\begin{abstract}

Federated Learning (FL) enables collaborative model training while preserving privacy by allowing clients (e.g., smartphones) to share model updates instead of raw data. Pervasive computing environments (e.g., for Human Activity Recognition - HAR), which we focus on in this paper, are characterised by resource-constrained end devices, streaming sensor data and intermittent user participation. Also, variations in user behaviour, common in HAR environments, often result in non-stationary data distributions. As such, existing FL approaches encounter challenges in such HAR environments due to their different underpinning assumptions to HAR environments. The combined effect of HAR environment characteristics, viz. heterogeneous data and intermittent participation, can lead to a severe problem called \emph{catastrophic forgetting} (CF).
Unlike Continuous Learning (CL), which addresses CF using shared memory and replay mechanisms, FL’s privacy constraints prohibit such strategies. To tackle CF in HAR environments, we propose \emph{FlexFed}, a novel FL approach that prioritizes data retention for efficient memory use and that dynamically adjusts offline training frequency based on distribution shifts, client capability and offline duration. Moreover, to better quantify CF in FL, we propose a more suitable metric that accounts for under-represented data, thereby providing more accurate evaluations. For FlexFed evaluation, we develop a realistic HAR-based framework to evaluate it against various approaches, including FedAvg, MIFA, and REFL. This framework simulates data streaming by capturing dynamic distributions, data imbalances and varying device availabilities, and precisely capture levels of forgetting.  Experimental results demonstrate FlexFed’s superiority in mitigating CF, improving FL efficiency by 10-15\% and achieving faster, more stable convergence, particularly for infrequent or under-represented data.
\end{abstract}


\section{Introduction}

Federated Learning (FL) \cite{mcmahan2017communication,konevcny2016federated} is a novel distributed Machine Learning (ML) training approach that allows numerous edge devices to collaboratively train an ML model while preserving data privacy. Recent FL implementations use FedAvg\cite{mcmahan2017communication} algorithm for client coordination.  In the standard FedAvg framework~\cite{mcmahan2017communication}, clients first receive the latest model parameters from the FL server. After conducting multiple local training iterations, they send the updated parameters back to the server for synchronization. This "communication round" repeats until the model converges or the termination condition is met.

Despite the growing popularity of FL in distributed ML, low training efficiency remains a significant barrier to practical deployment~\cite{lv2024fedca}. Most research in this domain operates under unrealistic assumption, that data remains static and accessible throughout the entire training process or participating clients consistently available. However, these assumptions fail to account for the challenges of real-world computing environments, such as pervasive computing, where resource-constrained devices (like smartphones or IoT devices) are owned by rational and independent users. As a result, data distribution is dynamic, device availability is unpredictable and network conditions fluctuate.  These challenges significantly affect FL performance, making it difficult to ensure efficient and reliable training in practical scenarios.

\begin{figure}[h]
    \centering
    \begin{subfigure}[b]{0.55\linewidth}
        \centering
        \includegraphics[width=\linewidth]{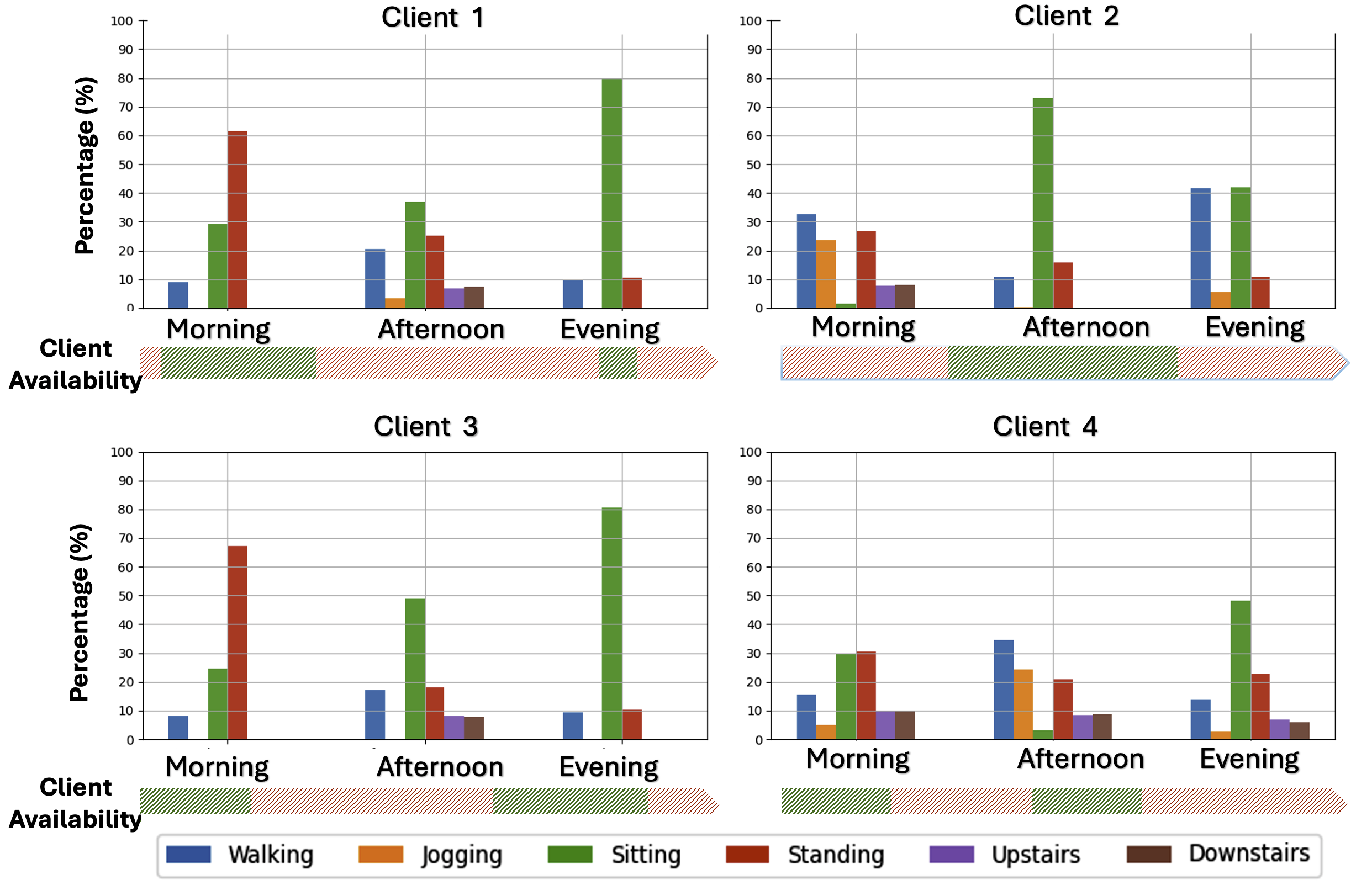}
        \caption{\centering Distribution of HAR Activities and Client Availability Throughout the Day.}
        \label{fig:Data_Distribution}
    \end{subfigure}
    \hfill
    \begin{subfigure}[b]{0.4\linewidth}
        \centering
        \includegraphics[width=\linewidth]{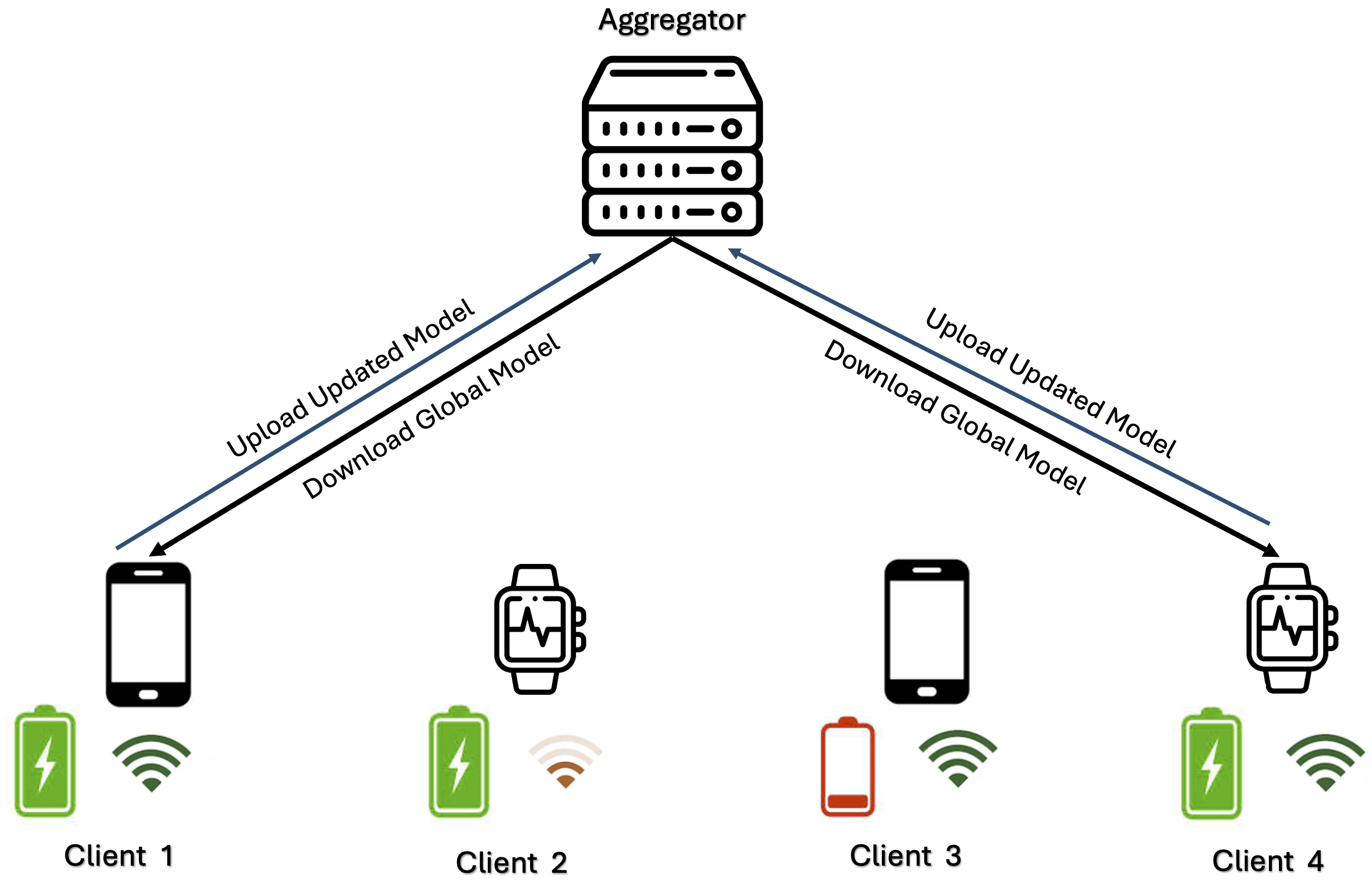}
        \caption{\centering Example Scenario of Training Round.}
        \label{fig:har}
    \end{subfigure}
   \caption{Overview of Human Activity Recognition (HAR) in a FL Setting.}
    \label{fig:HAR_System}
\end{figure}

To effectively deploy FL in pervasive computing environments, such as Human Activity Recognition (HAR), it is essential to identify key characteristics and constraints that influence performance, Figure~\ref{fig:HAR_System}. To explore this further, these characteristics can be classified as follows: (1) \textbf{Non-IID Data Distribution}: the data collected by individual clients exhibit significant variations in characteristics, patterns or distributions~\cite{kairouz2021advances}, shown in Figure~\ref{fig:Data_Distribution}. 
(2) \textbf{Streaming Data}: each client continuously acquires data from different sources, like sensors, meters or user interactions, these data usually follows a long-term label distribution.  (3) \textbf{Limited storage capacity}: The clients frequently have limited storage space. Therefore, typically, only the most recent data is kept, with previous data purged in a FIFO fashion, increasing the data dynamism. This exacerbates the mismatch between the long-term data distribution, which represents the client’s data over time and the short-term and time-varying data currently available in memory, as illustrated in Figure~\ref{fig:Data_Distribution}.
(4) \textbf{Intermittent availability}: The clients cannot consistently participate in all training rounds due to the participation requirements imposed, i.e., the device needs to be available\footnote{A client is available  when it is idle, connected to power and has the required bandwidth, e.g.,~\cite{abdelmoniem2023refl}, so training does not interfere with device performance.} ~\cite{abdelmoniem2023refl}.  Therefore, the client may become unavailable due to battery depletion or could move to an area with limited or no connectivity, as shown in Figure~\ref{fig:har}.

The aforementioned characteristics inevitably lead to three levels of data heterogeneity during the training rounds: \textbf{(i) intra-round}, arising from diverse client data distributions within a single training round; \textbf{(ii) inter-round}, caused by varying available client participation across rounds; and \textbf{(iii) intra-client}, shaped by client memory constraint,  leading to an inability to retain all the data until it is available to participate in an FL round. Together, these data heterogeneity factors lead to \textbf{catastrophic forgetting (CF)}, a well-known challenge in the continual learning domain \cite{chen2022lifelong},  where the globally trained model struggles to retain knowledge from earlier training data while adapting to the evolving data distributions provided by currently available training data \cite{venkatesha2022addressing,aljahdali2024flashback}.

Several approaches have been proposed to address CF in FL, such as regularization terms~\cite{li2020federated}, memory of gradients/updates~\cite{gu2021fast,luo2023gradma}, Synthetic pseudo-data~\cite{xu2022acceleration,qi2023better}, Knowledge Distillation~\cite{aljahdali2024flashback,li2019fedmd} dynamic aggregation and parameter optimization strategies~\cite{liu2023dynamite,jin2021budget,hong2022weighted}.
While these methods have their merits, they fail to comprehensively address CF while accounting for all the aforementioned challenges simultaneously without undermining the privacy guarantees of FL.  Another promising direction involves adopting memory management, such as selective caching~\cite{wang2023local,marfoq2023federated} and data valuation-driven selection~\cite{wei2024fedds,gong2023store}, by retaining informative samples and discarding less valuable ones. Despite efforts to preserve valuable patterns in training data, the inevitable intermittent client availability, strict memory limits and continuous data influx still prevent a   complete elimination of CF, which requires more investigation.

In fact, a core limitation across most of existing FL frameworks is the predominantly passive role assigned to clients, restricted to centrally controlled, binary participate-or-idle states contingent on stable  connectivity, idleness and battery condition~\cite{abdelmoniem2023refl}. Client contributions are severely limited by such strict participation requirements, particularly in pervasive computing contexts, which are characterised by varying resources, sporadic connectivity and changing data distribution. To bridge these critical gaps, we propose \emph{\textbf{FlexFed}}, an FL framework designed to proactively utilize client-side capabilities through intelligent local training and adaptive, class-wise data management to reducing the CF and significantly advancing FL performance in pervasive computing scenarios. We specifically investigate FlexFed on the well-known WISDM dataset from HAR given that CF is a significant and common challenge in real-world HAR deployments, as HAR scenarios naturally involve highly dynamic and continuously evolving data streams and clients frequently encounter intermittent connectivity and limited resources. 

\section{Background}
\subsection{Federated Learning (FL)}
Since its proposal by~\cite{mcmahan2017communication}, FL has become a widely adopted framework for large-scale distributed learning, enabling various ML tasks while preserving privacy~\cite{aljahdali2024flashback}. It consists of two primary entities: a central server \(PS\) and a set of clients \(\mathcal{K}\).  Each client \( k \in \mathcal{K} \) maintains a local dataset \( \mathcal{T}_k \), consisting of samples \( \{ z_{i} = (x_{i,k}, y_{i,k}) \in X \times Y : i \in \mathcal{T}_k\} \), where \( x_{i,k} \) represents the feature of the \( i \)-th sample in feature space \( X \) and  \( y_{i,k} \) is its corresponding label in label space \( Y \).  In a typical FL algorithm, such as FedAvg~\cite{mcmahan2017communication}, the primary objective is to solve an empirical risk minimization problem such as:
\small
\begin{equation}
\min_{\theta}\mathcal{L}(\theta) = \min_{\theta} \frac{1}{|\mathcal{K}|} \sum_{k \in \mathcal{K}} \mathcal{L}_{k}(\theta)
\end{equation}
\normalsize
, where \( \theta \) denotes the model parameters to be optimized:
\small
\begin{equation}
    \mathcal{L}_{k}(\theta) = \frac{1}{|\mathcal{T}_{k}|} \sum_{z \in \mathcal{T}_{k}} \mathcal{L}(\theta; z) 
\end{equation}
\normalsize
   is the empirical loss function at client \( k \), 
    and \( \mathcal{L}(\theta; z) \) is the loss function evaluated at model \( \theta\) and data sample \( z \).  Training proceeds over a set of rounds, denoted by \(\mathcal{R}\). In each round \(r \in \mathcal{R}\), the server sends the global model \(\theta^{r-1}\) to selected group of available clients \(\mathcal{S}^r \subseteq \mathcal{K}\) 
, where $\mathcal{S}^r$ is a subset of available clients selected for participation. Each client \( k \in \mathcal{S}^r \) updates its local model \( \theta^{r}_k \) using its dataset \( \mathcal{T}_k \), then sends the updated model to the \(PS\), which aggregates the received updates—typically by averaging—to generate a new global model \( \theta^{r} \). This procedure is repeated until a termination criterion is fulfilled.

\subsection{Forgetting in FL}

Initial theoretical analyses of FL focused on IID data and full client participation~\cite{stich2018local}, resulting in linear convergence under optimal conditions. However, these assumptions are unrealistic in HAR environments. Several works that focused on the use of non-IID data and partial client participation have demonstrated that these challenges hinder convergence guarantees~\cite{yang2021achieving}.  These shifts present significant challenges, especially when the evolving training data results in concept drifts, described as the continuous changing of a global model's parameters as it undergoes retraining on fresh data from selected clients in each round, frequently emphasising recent inputs. This adaptation may result in the loss of previously acquired patterns, a phenomenon known as \emph{catastrophic forgetting }\cite{aljahdali2024flashback,li2019fedmd}.
Mathematically CF occurs if:
\small
 \begin{equation}
  \mathcal{L}(\theta^{r+1}; \mathcal{T}_k) > \mathcal{L}(\theta^{r}; \mathcal{T}_k)
 \end{equation}
\normalsize
This phenomenon negatively impacts under-represented data, including infrequent labels, significantly decreasing their impact on the model. The swift decline in representation happened when clients carrying these labels are frequently unavailable or when participants lack sufficient memory to retain such data.

In Continual Learning (CL), CF is usually measured through the Backward Transfer (BwT) metric~\cite{chaudhry2018riemannian}. This metric was  adapted to suit FL in ~\cite{lee2022preservation} as follows:
\small
\begin{equation}
F = \frac{1}{|C|} \sum_{c=1}^{|C|}\max_{r \in \{1, \ldots, \mathcal{R}-1\}} (Acc_{c}^{r} - Acc_{c}^{R})
\label{ref:Backward_Transfer}
\end{equation}
\normalsize
where \(| C |\) represents the number of labels, \( Acc_{c}^{r} \) denotes the global model's accuracy on class \( c \) at round \( r \) and  \( Acc_{c}^{R} \) is the accuracy achieved by the global model in the final round \( R \).

\subsection{Role of Intermittent Availability and On-Device Storage Constraints in Forgetting During Federated Training}

The inconsistent availability and constrained resources hinder the continuity of learning equality across all labels, which could result in models failing to accurately represent a client's complete data history. This often exacerbates model drift and increases the risk of catastrophic forgetting. As illustrated in Figure~\ref{fig:Data_Distribution}, if client 1 is absent during an afternoon training round, due to unavailability, it misses the opportunity to learn from the unique data available at that time. Over time, these missed opportunities can cause the local model to become outdated and the client’s long-term data distribution becomes underrepresented in the local model (subsequently in global model), especially if its data distribution differs from those of other clients.

\begin{figure}[htbp]
    \centering
    \begin{subfigure}{0.48\linewidth}
        \centering
        \includegraphics[width=\linewidth]{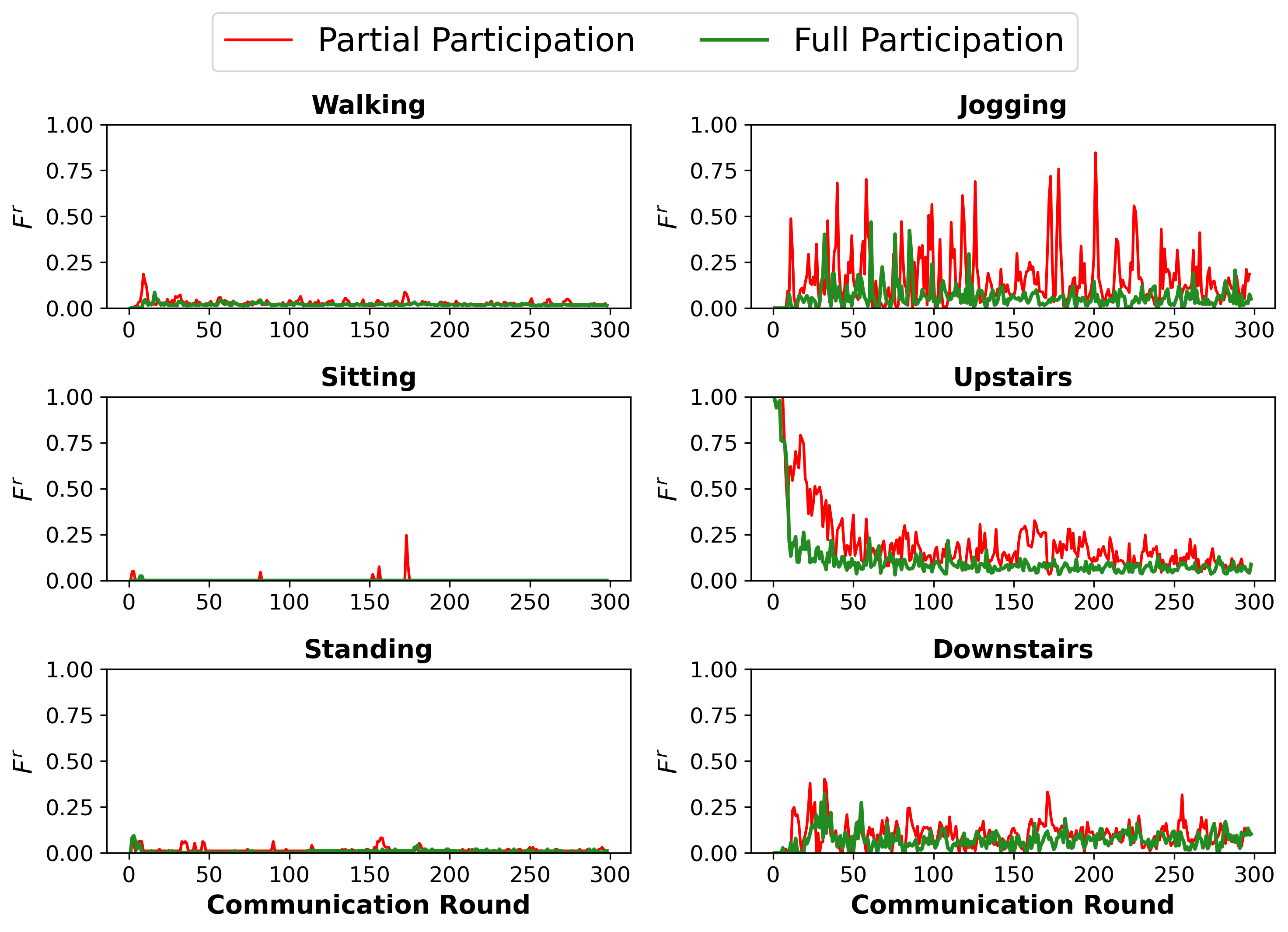}
        \caption{Forgetting levels ($F^r$) per class driven by intermittent participation.}
        \label{fig:intermittent_availability}
    \end{subfigure}
    \hfill
    \begin{subfigure}{0.48\linewidth}
        \centering
        \includegraphics[width=\linewidth]{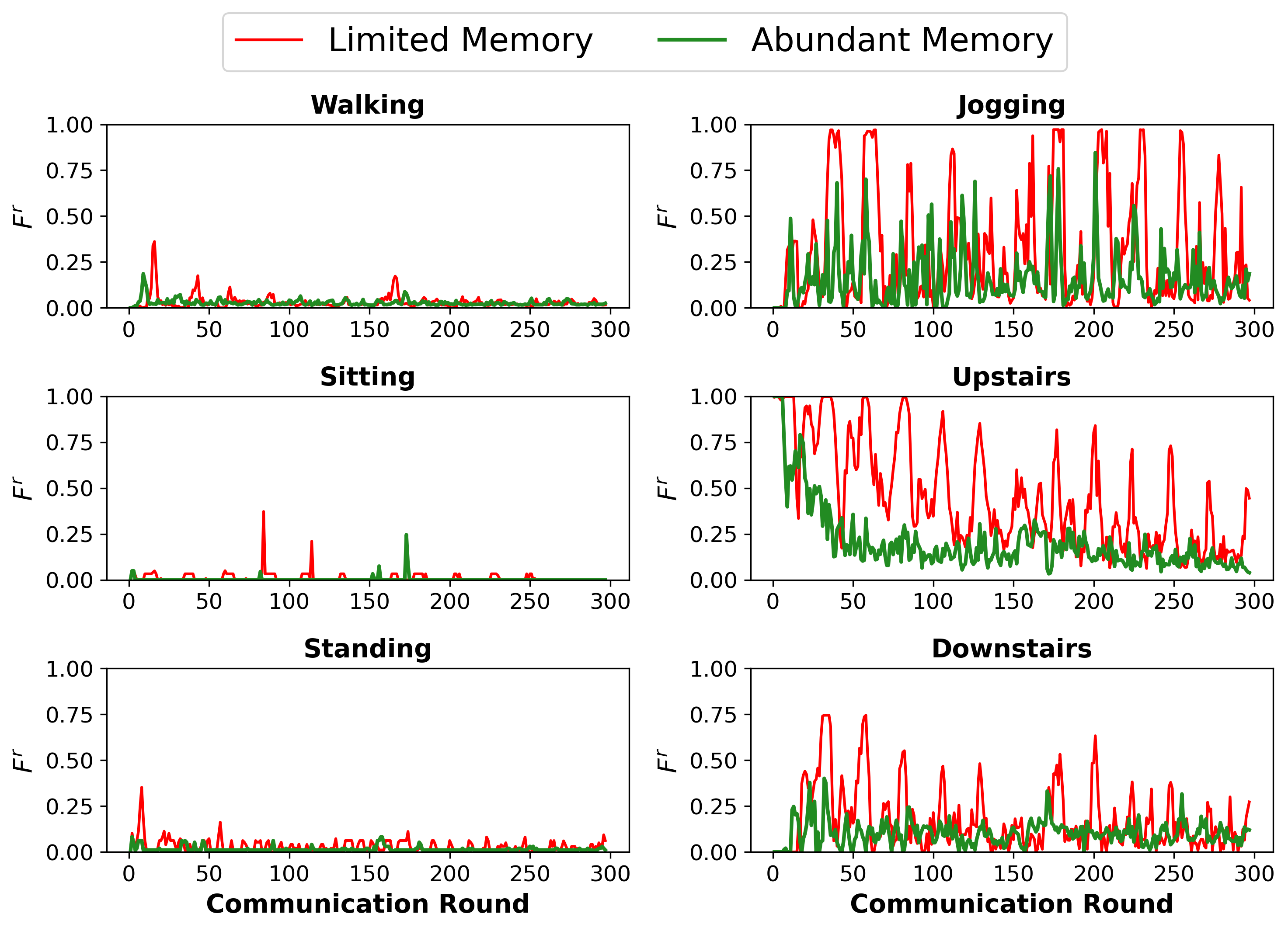}
        \caption{Forgetting levels ($F^r$) per class driven by memory constraints.}
        \label{fig:memory_constraints}
    \end{subfigure}
    \caption{Impact of (a) Intermittent Availability and (b) Memory Constraints on FL Performance.}
    \label{fig:memory_vs_availability}
\end{figure}

To assess the impact of clients (un)availability and storage limitations on federated training performance, we conducted experiments on the HAR dataset (details in the experiment setup section) and calculated the forgetting value, Figure~\ref{fig:memory_vs_availability}. To accurately measure forgetting and avoid the knowledge replacement dilemma—where improvements in one class obscure the declines in another, thereby masking forgetfulness and  computed the average forgetting across all clients — The forgetting level was measured using an adapted version of Equation \ref{ref:Backward_Transfer}. For example, if the model's accuracy for "upstairs" decreases but is offset by an increase in accuracy for "walking," the overall performance may seem stable or improved.  To better manage forgetting, we compare the model's accuracy between the current and previous rounds to detect any performance declines. Thus, we propose an updated metric for forgetting, as an alternative to that proposed in~\cite{aljahdali2024flashback}, as follows:
\small
\begin{equation}
F^{r} = \frac{1}{|C|} \sum_{c=1}^{C} \frac{1}{\mathcal{|K|}} \sum_{k=1}^{\mathcal{K}} \min(0, (Acc_{k,
c}^{r} - \max_{i \leq r} Acc_{k,c}^{i}))
\label{BtW_equation}
\end{equation}
\normalsize
where \( F^r \) represents the forgetting metric for the current round \( r \), \( \mathcal{|K|} \) denotes the total number of clients and  \( |C| \) represents the total number of labels, \(Acc_{k,c}^{r}\) is the accuracy of client \(k\) per class \(c\) in the current round \(r\) and \(\max_{i \leq r} Acc_{k,c}^{i}\) is the highest accuracy for that class $c$ achieved by client \(k\) up to and including the current round \(r\). The \(\min(0, \cdot)\) function ensures that only negative differences (indicating a drop in accuracy) are considered, thus capturing the effect of catastrophic forgetting.  The adjustment is based on our assumption that each client will use their own test data to assess the global model in order to figure out if their previously gained knowledge has degraded or not for every class. After that, we ensure that all Labels  are assigned equal value, regardless of their frequency in the client's test data. 

\textbf{Intermittent Availability:} To properly examine the influence of participation rates on model performance, we assume that clients possess static data, which isolates the impacts of memory restrictions. Once we included variations in client availability patterns (using IMA dataset), we observed that client participation rates have a considerable impact on forgetting. This impact is especially pronounced for underrepresented labels, as illustrated in Figure~\ref{fig:intermittent_availability}.

\textbf{Memory Constraints:} On the other hand, to study the impact of memory constraints, we introduce this factor as an additional variable in our analysis. Figure~\ref{fig:memory_constraints} illustrates the differences in forgetting values across labels due to these memory constraints, especially for underrepresented labels.

\section{Related Work}
\textbf{Data Heterogeneity in FL.}  FedAvg~\cite{mcmahan2017communication} is effective when the training data is homogeneous (Independent and Identically Distributed, IID) across different clients. However, when the data is heterogeneous and diverse (Non-IID), FedAvg struggles to train a robust global model, resulting in biased or suboptimal performance~\cite{chen2018inferring}.  
There are several algorithm designed to handle the challenges posed by non-IID data in federated learning.  One direction focuses on local-side modification,  where the regularisation of the local models' divergence from the global model, e.g., FedProx~\cite{li2020federated}, SCAFFOLD~\cite{karimireddy2020scaffold}, Moon~\cite{li2021model} and FedDyn~\cite{acar2021federated}. The other direction performs changes on the server-side, which boosts the effectiveness of the aggregation of local models within the server, e.g.,FedMA~\cite{wang2020federated}, Fedbe~\cite{chen2020fedbe}. 

\textbf{Forgetting in FL: } CF caused by the loss of previously acquired knowledge during training, presents a considerable challenge in FL. Recently, a variety of strategies have been proposed to address this issue. FedCurv~\cite{shoham2019overcoming} exemplifies an initial attempt  by considering each client as a distinct task and controls local parameter modifications. FedNTD~\cite{zhao2018federated} and Flashback~\cite{aljahdali2024flashback} leverages distillation strategy, sharing globally representative synthetic samples across tasks to retain knowledge. FedReg~\cite{xu2022acceleration} proposed  synthetic pseudo-data for regularization during local updates, to balancing global and local perspectives.  Methods such as MIFA~\cite{gu2021fast} and GradMA~\cite{luo2023gradma} focus on the utilization of stored updates; however, both increase memory and computational demands, especially at scale.In the context of streaming data, ~\cite{wang2023local,marfoq2023federated,wei2024fedds,gong2023store},  emphasize memory management to alleviate forgetting. Challenges such as partial participation, label imbalance, privacy risks and  computational overhead impede the effectiveness of these approaches. This works extends previous research by addressing dynamic data, intermittent availability and  storage limitations to enhance forgetting mitigation in FL.

\section{Forgetting Formalisation}
\label{sect:pbm-stmt}
We investigate FL in HAR environments where clients \( k \in \mathcal{K} \) continuously collect data from diverse sources but face challenges as previously described.
The optimization objective is to learn an optimal global model $\theta$ which can generalize well to the dataset of all the $|\mathcal{K}|$ clients $\{\mathcal{T}_k\}_{k=1}^\mathcal{|K|}$:
\small 
\begin{equation}
\theta = \arg \min_\theta \sum_{k=1}^{|\mathcal{K}|} \frac{n_k}{n_S} L_k(\theta; \mathcal{T}_k)
\end{equation}
\normalsize
where $\theta \in \mathbb{R}^d$ encodes the parameters of the global model and $L_k$ represents the loss incurred by client $k$ when fitting the model $\theta$ to its local dataset $\mathcal{T}_k$, $
L_k := \mathbb{E}_{(x,y) \sim \mathcal{T}_k}[L_k(\theta; \mathcal{T}_k)]$, 
  \(n_k\) is the number of samples that are held by client \(k\) and 
\(n_S\) denotes the total number of samples held by all selected clients \(S\).

Given a fixed memory of size \( m \), each client \( k \) maintains a dataset \( \mathcal{M}_k^r \subset \bigcup_{i=0}^{r} \mathcal{T}_{k}^{i} \) with \( |\mathcal{M}_k^r| \leq m \). When the memory is full and newer data is available, a FIFO replacement strategy is usually used.
Memory limitations and the constant influx of new data necessitate the use of this strategy~\cite{gong2023store}. Due to memory size limitations, the dataset of client \( k \) is represented as follows: 
$ \mathcal{T}_k = \{\mathcal{T}_{k}^{1}, \mathcal{T}_{k}^{2}, \ldots, \mathcal{T}_{k}^{r}, \ldots, \mathcal{T}_{k}^{R} \}, \quad \mathcal{T}_{k}^{r} = ( D_{k}^{r}) $, where \(\mathcal{T}_k\) is the collection of datasets over total rounds \(R\) and \(\mathcal{T}_{k}^{r}\) represents the actual dataset at round \( r \) store in memory \(m\) consisting of data points \( D_{k}^{r} \), such that  $D_{k}^{r} = \{X_{k}^{r}, Y_{k}^{r}\}$ and
the\(X_{k}^{r}\) represents the input data for the 
k-th client and \(Y_{k}^{r}\) represents the corresponding labels, the labels belong to \(C\). Since the data distribution is Non-IID: $
C_{k}^{i} \neq C_{k}^{j} \quad \text{for} \quad i, j \in \mathbb{R} \quad \text{and} \quad i \neq j$, where $C_k^i$ is the number of classes at client $k$ in round $i$. Moreover, we represent client availability in round \( r \) as:
\small
\[
\delta_k^r =
\begin{cases}
1 & \text{if client } k \text{ is available in round } r, \\
0 & \text{if client } k \text{ is not available in round } r.
\end{cases}
\]
\normalsize
Here, \(\delta_k^r\) indicates the availability of client \( k \) in round \( r \). A certain proportion of clients must be available for training in round \( r \) to proceed. Let \( \delta^r \) denote the set of all available clients in round \( r \).Then, the condition to complete round \(r\) successfully is, $ 
\frac{|\delta^r|}{|\mathcal{K}|} \geq \beta, \quad \forall r \in R
$, where~\(\beta \) is the minimal proportion of clients needed for each round to run effectively. Due to memory constraints and inconsistent availability, training a global model through multiple server-client communication on accumulated client-side tasks (\(\mathcal{T}^1, \mathcal{T}^2, \mathcal{T}^3, \) etc.) may need more time to converge for all tasks.  As data from previous tasks will be mostly unavailable for training, some of previously obtained knowledge will be lost. Therefore, the global model may lack knowledge of the full data distribution over tasks. 

Therefore, given the global model parameters received from the previous round, denoted by \(\theta^{r-1}\) and the new task dataset is \(\mathcal{T}^r = \bigcup_{k=1}^{\mathcal{K}} \mathcal{T}_k^r\), where \(\mathcal{T}_k^r\) comprises the recently acquired data at each client, our objective is to train a global model that minimizes the loss on both the new task \(\mathcal{T}^r\) and the previous tasks \(\{\mathcal{T}^1, \ldots, \mathcal{T}^{r-1}\}\). Therefore, the optimization objective is to minimize the losses of \(\mathcal{K}\) clients on all local tasks up to round \(r\)   could be defined as follows:
\small
\begin{equation}
\theta^r = \arg\min_{\theta} \sum_{k=1}^{K} \frac{n_k}{n_S} \sum_{i=1}^{r} \mathcal{L}(\mathcal{T}_k^i; \theta)
\end{equation}
\normalsize

To ensure the efficiency of federated training, we endeavour to have the global model at task $\mathcal{T}^r$ to have a loss that is not greater than the loss of the model at round $r - 1$ on previous tasks, as captured in the following equation:
\small
\begin{equation}
\sum_{i=1}^{r-1} \mathcal{L}(\mathcal{T}^i; \theta^r) \leq \sum_{i=1}^{r-1} \mathcal{L}(\mathcal{T}^i; \theta^{r-1}).
\label{equation:difference}
\end{equation}
\normalsize

\section{FlexFed: A Methodology to Mitigate Forgetting}
\label{sect:metho}

 FlexFed introduces two key enhancements for robust FL: (i) Offline Client Training, enabling clients to independently initiate model updates locally during periods of poor connectivity. When a stable connection is restored, these locally refined models are synchronized with the server, maintaining training continuity.  (ii) Performance-Adaptive Memory Management, which intelligently selects data subsets based on the performance of local model to reduce memory overhead and ensure efficient training.  


\subsection{Offline Client Learning via Opportunistic Resource-Aware Scheduling}

As previously stated, existing FL strategies assume that a client initiates local model training only when selected to participate in a training round, necessitating three conditions for availability: (i) a WiFi connection, (ii) the device being idle~\cite{abdelmoniem2023refl} and (iii) having sufficient battery capacity. If these prerequisites are not met, the client's local model will remain unchanged until it can participate in the training process. Our goal therefore is to develop a training strategy so that clients can train locally during offline available times\footnote{Offline available times refer to periods when the client is idle and has power to conduct local training but lacks a stable connection to deliver the model to the server.}, when a stable connection is lacking but the other two conditions are met.  Increasing the frequency of local updates facilitates the processing of additional training samples, thereby improving the accuracy of the local model and a more representative data distribution than the old fashion.  This approach require trade-offs, such as performance gain, considering computational costs and the possibility of higher divergence between local and global models, which may slowing convergence. To address these challenges, we assume that clients make intelligent locally decisions—such as forecasting idle times, charging periods and data stream rates —by employing time-series forecasting techniques, to figure out the optimal number of offline training sessions.


\subsection{Performance-Adaptive Memory Allocation}

Given the client's limited memory capacity \(m\), we propose a dynamic memory management approach based on performance, i.e., memory is allocated for additional data based on the client's model performance \(\alpha_{k}\). In our proposed method, the proportion of memory dedicated to storing additional data \(\zeta_k^r\) after round \(r\) is inversely proportional to the client's performance \(\alpha_{k}\). Specifically, clients demonstrating higher local model performance allocate less memory to retain data from previous tasks, whereas clients with lower local model performance allocate more memory to improve learning from low representation data. Formally,  during initial task, the additional memory is initialized as empty:
\small
\[
|\zeta_k| = 0
\]
\normalsize
Then, the memory allocation for additional data at client \(k\) is adjusted as follows:
\small
\[
|\zeta_k| = m \times (1 - \alpha_k)
\]
\normalsize
The parameter \( \alpha_{k} \in [0, 1] \) is determined based on the performance of the received model \( \theta^{r-1} \) on the test data \( D_k^{\text{test}} \). A lower value of \( \alpha_{k} \) indicates better performance of the previous model, resulting in a smaller portion of old data \(\zeta_k\) being retained. Conversely, a higher value of \( \alpha_{k} \) indicates poorer performance, resulting in a larger portion of old data \(\zeta_k\) being retained.

The set \(\zeta_k^r\) consists exclusively of samples from infrequent classes \(C'\). Thus, the additional data \(\zeta_k^r\) stored after round \(r\) given by:
\small
\[
\zeta_k^r \subseteq \bigcup_{d=1}^{r-1} \mathcal{T}_{k}^{d}(C')
\]
\normalsize

Here, \(\mathcal{T}_{k}^{d}(C')\) is defined as the set of samples from task \(d\) belonging to infrequent classes \(C'\):
\small
\[
\mathcal{T}_{k}^{d}(C') = \{ x \in \mathcal{T}_{k}^{d} \mid \text{class}(x) \in C' \}
\]
\normalsize

The training data for client \(k\) in round \(r\), denoted as \(\mathcal{T}_{k}^{r}\), is defined as:
\small
\[
\mathcal{T}_{k}^{r} = \zeta_k^r \cup \mathcal{T}_{k}
\]
\normalsize
where \(\mathcal{T}_{k}\) represents the captured data in memory.

\subsection{FlexFed Training Process Workflow}
The training  occurs over multiple global round between the server and the clients, as following: {\textbf{1) Initialisation Phase:}}
The model parameters  \( \theta^0 \) are initialized, randomly or with a public dataset and set training parameters, e.g., round duration \(\mathcal{L}\) and selection fraction \(\tau\) are configured.
 {\textbf{2) Advertising and Selection Phase:}}
The \(PS\) broadcasts the next round’s interval and task requirements, then selects a subset \( S \) of \( \tau \cdot K \) available clients from \( N \), excluding those chosen in the previous round to prioritize others. {\textbf{3) Distribution and Training phase:}} The server distributes the model and each client then performs local training using $
\theta_{k}^{r+1} = \theta_{k}^{r} - \eta \nabla L(\theta_{k}^{r}; \mathcal{T}_{k}^{r})$, where $\theta_{k}^{r}$ are client $k$'s model parameters at round $r$, $\eta$ is the learning rate, $L$ is the loss function and $\mathcal{T}_{k}^{r}$ represents the local data. After training, clients evaluate the updated model against the stored version. If performance improves—measured by metrics such as accuracy the updated model is sent back to the server; otherwise, the stored model is retained and returned to the server.

The performance comparison process can be represented as:

\[
\begin{aligned}
& \text{If } (\mathcal{P}(\theta_{k}^{r+1}; D_k^{\text{test}}) \geq \mathcal{P}(\theta_{k}^{\text{stored}}; D_k^{\text{test}})) \text{ then} \\
& \quad \text{send } \theta_{k}^{r+1} \text{ to the server and update } \theta_{k}^{\text{stored}} = \theta_{k}^{r+1}; \\
& \text{Else:} \\
& \quad \text{send } \theta_{k}^{\text{stored}} \text{ as } \theta_{k}^{r+1} \text{ to the server and discard } \theta_{k}^{r+1};
\end{aligned}
\]

\noindent
where $\mathcal{P}(\theta; D_k^{\text{test}})$ is the performance of the model $\theta$ on the test data $D_k^{\text{test}}$,  e.g., accuracy value and $\theta_{k}^{\text{stored}}$ is the stored model at client $k$.

{\textbf{4) Global Aggregation: }}
The server performs Staleness-based Aggregation \cite{abdelmoniem2021resource} for receiving updates. The aggregation process can be formalized as:
\small
\begin{equation}
\theta^{r+1} = \frac{1}{\mathcal{K}} \sum_{k=1}^{\mathcal{K}} \gamma_{k} \theta^{r+1}_{k}
\end{equation}
\normalsize
where, $\gamma_{k}$ is the scaling factor for the stale updates from client $k$, which depends on the number of late rounds.

\begin{figure}
    \centering
    \includegraphics[width= 0.7\linewidth]{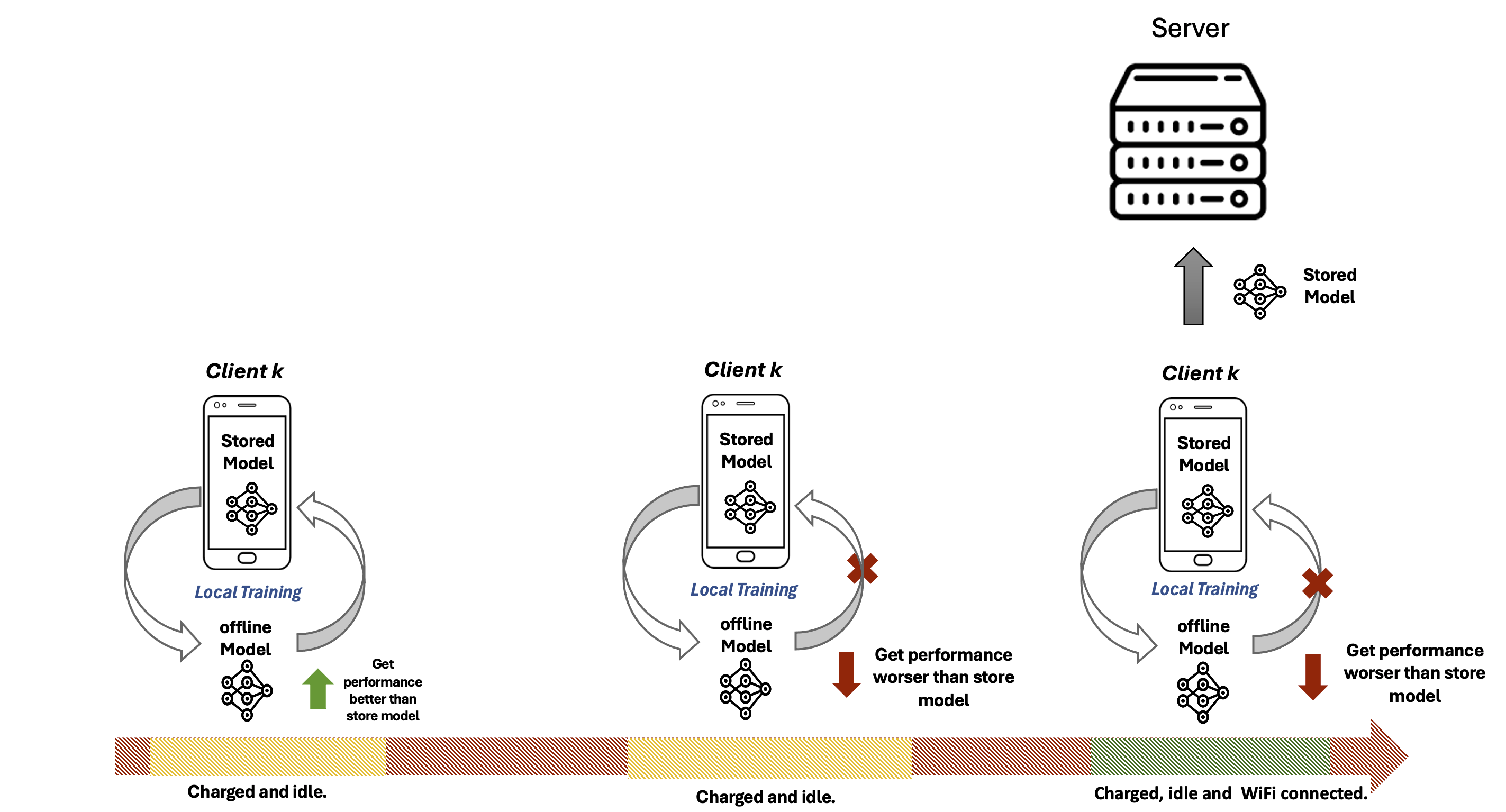}
    \caption{Procedures of Client k's  Training Throughout a Day.}
    \label{fig:Training}
\end{figure}

{\textbf{5) Offline Training :}}
If a client is idle and powered (with no connection to \(PS\)), it is considered eligible for offline training, where it continues training locally. After each offline training session, the client compares the performance of the updated local model with the stored model. If the performance improves, the updated model is stored; otherwise, it is discarded. Performance \(\mathcal{P}\) can be measured using metrics such as accuracy on test data, as illustrated in Fig.~\ref{fig:Training}. The offline training update can be formalized as:
\begin{equation}
\theta_{k}^{\text{offline}} = \theta_{k}^{stored} - \eta \nabla L(\theta_{k}^{stored}; \mathcal{T}_{k}^{r})
\end{equation}
Then, \( \theta_{k}^{\text{stored}}\) is updated as follows:
\small
\[
\begin{aligned}
& \text{If } (\mathcal{P}(\theta_{k}^{\text{offline}};  D_k^{\text{test}}) \geq \mathcal{P}(\theta_{k}^{\text{stored}}; D_k^{\text{test}})) \\
& \text{Then: } \theta_{k}^{\text{stored}} := \theta_{k}^{\text{offline}}; \\
& \text{Else: discard } \theta_{k}^{\text{offline}};
\end{aligned}
\]
\normalsize

{\textbf{6) Client's Data Management:}} To update the amount of \( \zeta_k^r \), once client \(k\) selects to participate in round \(r\), it updates its \(\alpha_{k}\), as $
\alpha_{k} = \mathcal{P}(\theta^{r} ,D_{k}^{\text{test}})
$, where, \(\alpha_{k}\) is determined based on the performance of the latest received model from the server \(\theta^{r}\) on the client test data \(D_{k}^{\text{test}}\). The aim is to reduce stored data if the client's performance is high and not affected by forgetting, thereby optimizing memory utilization. Otherwise, it keeps stores data to enhance training.

\section{Experimental Setup}
\label{sect:expt-setup}

We run experiments using the FedScale framework  \cite{lai2022fedscale,abdelmoniem2021resource} and an NVIDIA GPU cluster to interleave the training of the simulated clients using PyTorch v3.8.17. The hyper-parameters for training are configured as follows: $\mathcal{K}=100$, $R=300$ rounds, maximum round length $L=300$, local iterations $E=10$, batch size $B=32$ and  learning rate $\gamma=0.005$. We evaluate the performance of FlexFed using six popular HAR models in our experiments: CNN, ResNet, ShuffleNet, Swin, ViT and  MobileNet. The IMA dataset \cite{yang2022flash} is utilised to reflect the heterogeneity in device capabilities and  availability among clients.

\begin{figure}[H]
    \centering
    \includegraphics[scale=0.23]{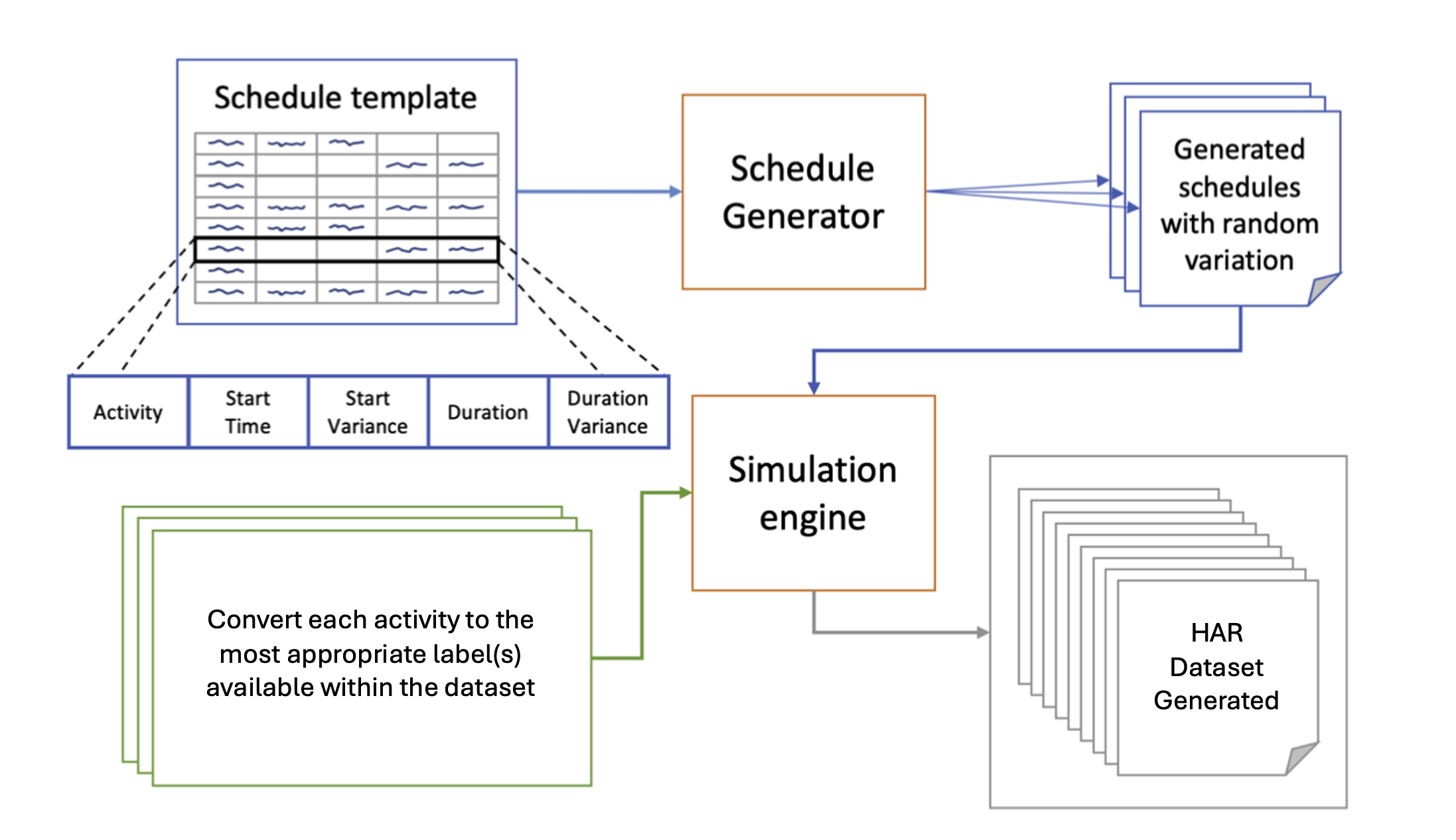}
    \caption{Generation of HAR Datasets \cite{idrees2023framework}.}
    \label{fig:Generate_HAR}
\end{figure}

Most datasets used in FL research, such as CIFAR-10~\cite{krizhevsky2009learning} and MNIST~\cite{lecun1998gradient}, are static in nature, meaning the data is pre-collected and distributed to clients before training begins. Such static datasets fail to capture the temporal dynamics and evolving characteristics of real-world applications. Motivated by the limitations of existing datasets and the need to more accurately reflect data nature in real-world conditions, we focus on HAR and introduce a dynamic dataset generation process. HAR serves as an effective domain for studying temporal dynamics and forgetting not only due to its streaming sensor data and deployment on resource-constrained devices, but also because it encompasses diverse activity patterns, frequent class imbalances~\cite{leite2022resource,schiemer2023online}. Our methodology builds upon the WISDM~\cite{krishnan2008real} dataset (see Table~\ref{tab:dataset_statistics}), inspiration from the general  framework proposed in~\cite{idrees2023framework}, as illustrated in Figure~\ref{fig:Generate_HAR}. This approach enables the creation of evolving datasets during training, better capturing the non-stationary and heterogeneous nature of data in practical federated learning scenarios.  Due to page constraints, we focus on evaluating our approach using a single dataset; however, future work will extend these results to a broader range of HAR datasets.

\begin{table}[tbp]
    \centering
    \caption{\small Daily Routine Schedule Template Example with Time and Duration Variances.}
    \tiny
    \begin{tabular}{|c|c|c|c|c|c|}
        \hline
        & \textbf{Activity} & \textbf{Start time} & \textbf{ Start variance} & \textbf{Duration} & \textbf{Duration variance} \\
        \hline
        1 & Shower & 7:00 am & 00:20 & 00:30 & 00:05 \\
        \hline
        2 & Breakfast & 7:30 am & 00:15 & 00:20 & 00:05 \\
        \hline
        3 & Brush teeth & 7:50 am & 01:15 & 00:10 & 00:00 \\
        \hline
        4 & Transportation & 8:00 am & 00:30 & 01:00 & 00:25 \\
        \hline
        5 & Work & 9:00 am & 00:00 & 04:00 & 00:05 \\
        \hline
        6 & Lunch & 1:00 pm & 01:00 & 00:30 & 00:10 \\
        \hline
        7 & Work & 1:30 pm & 00:00 & 01:30 & 00:30 \\
        \hline
        8 & Watch TV & 4:30 pm & 00:15 & 01:30 & 00:25 \\
        \hline
        9 & Dinner & 6:00 pm & 00:20 & 00:45 & 00:15 \\
        \hline
    \end{tabular}
    \label{table:template_example}
\end{table}


Initially, the framework creates schedule templates for different client groups (e.g., elderly, students, employees) that define usual daily activity sequences with varying activity start/end time  and variability (see Table \ref{table:template_example}). Using these templates, the schedule generator produces diverse 14-day schedules for all clients. Afterwards, the converter associates each activity with its most relevant labels and the associated percentages, based on Table~\ref{tab:activity_tasks}. For example, the (work) activity is associated with 30\% sitting, 30\% standing, 20\% walking, 10\% upstairs and  10\% downstairs.

\begin{table}[h!]
\centering
\scriptsize
\caption{Statistics of WISDM and UCI HAR Datasets.}
\label{tab:dataset_statistics}
\begin{tabular}{@{}lcccc@{}}
\toprule
\textbf{Dataset} & \textbf{Classes} & \textbf{Samples} & \textbf{Data Type} & \textbf{Participants} \\
\midrule
WISDM~\cite{weiss2019wisdm} & 6 & \makecell{1,098,207 \\ samples} & \makecell{Accelerometer \\ and Gyroscope \\ readings} & 36 users \\

\bottomrule
\end{tabular}
\end{table}

\normalsize

\begin{table}[tbp]
\centering
  \caption{Activity vs the Most Relevant Label(s).}
\label{tab:activity_tasks}
\tiny
\begin{tabular}{|c|c|c|c|c|c|c|}
\hline
\textbf{Activity} & \multicolumn{6}{c|}{\textbf{label}}                \\ \hline

                  & \textbf{Sitting} & \textbf{Standing} & \textbf{Walking} & \textbf{Jogging} & \textbf{Upstairs} & \textbf{Downstairs} \\ \hline
Shower            & -                & 0.9               & 0.1              & -                & -                  & -                    \\ \hline
Breakfast         & 0.8              & 0.1               & 0.1              & -                & -                  & -                    \\ \hline
Brush teeth       & -             & 0.9               & 0.2                 & -                & -                  & -                    \\ \hline
Transportation    & 0.5              & -                 & 0.5              & -                & -                  & -                    \\ \hline
Work              & 0.3             & 0.3              & 0.2             & -                & 0.1                 & 0.1                    \\ \hline

At Park              & 0.2             & 0.1              & 0.6             & 0.1               & -                & -                   \\ \hline

At School              & 0.5             & 0.2              & 0.2             & -               & 0.1                & 0.1                   \\ \hline

Lunch             & 0.8              & 0.1               & 0.1                & -                & -                  & -                    \\ \hline
Watch TV          & 0.8              & 0.1               & 0.1                & -                & -                  & -                    \\ \hline
Study              & 0.8             & 0.1              & 0.1             & -               & -               & -                   \\ \hline

Workout           & -                & 0.1                 & 0.4              & 0.3              & 0.1                & 0.1                  \\ \hline
Dinner            & 0.8              & 0.1               & 0.1              & -                & -                  & -                    \\ \hline
\end{tabular}
\end{table}

After that, we utilize this schedule to distribute the  WISDM dataset \cite{krishnan2008real} across the clients.  To simulate limited memory in devices, a sliding window strategy \cite{xiao2021federated} is utilised.  We build a sequence of segments $ S= \{S_1, S_2,..., S_{Nseg}\}$ that act as the input dataset for local model $\omega$, with \(Nseg\) representing the total number of segments.  
Therefore, for any client, $S_k$ is made up of a $l$-long sequence, where $l$ is the length of the sequence, produced from data acquired by sensors, e.,g last 12 hours.

We have chosen two well-known approaches, {REFL}~\cite{abdelmoniem2023refl} and {MIFA}~\cite{gu2021fast} for comparisons. REFL manages intermittent connectivity by selecting available clients, while MIFA  retains recent updates. We evaluate the performances of the algorithms using the following metrics: (i) {Overall Performance:} Global model test accuracy  and loss, (ii) {Per Class Performance:} Test accuracy and loss for each class, (iii) {Overall Forgetting Rate:} Average difference between current and highest accuracy per round across all labels (the disparities between the current round's accuracy and the highest accuracy value achieved up to that round) and (iv) {Per Class Forgetting Rate:} Difference between the current and highest accuracy per round for each class.

\section{Results}
\label{sect:results}

\begin{figure}
    \centering
    \includegraphics[width=1\linewidth, height=4.5cm]{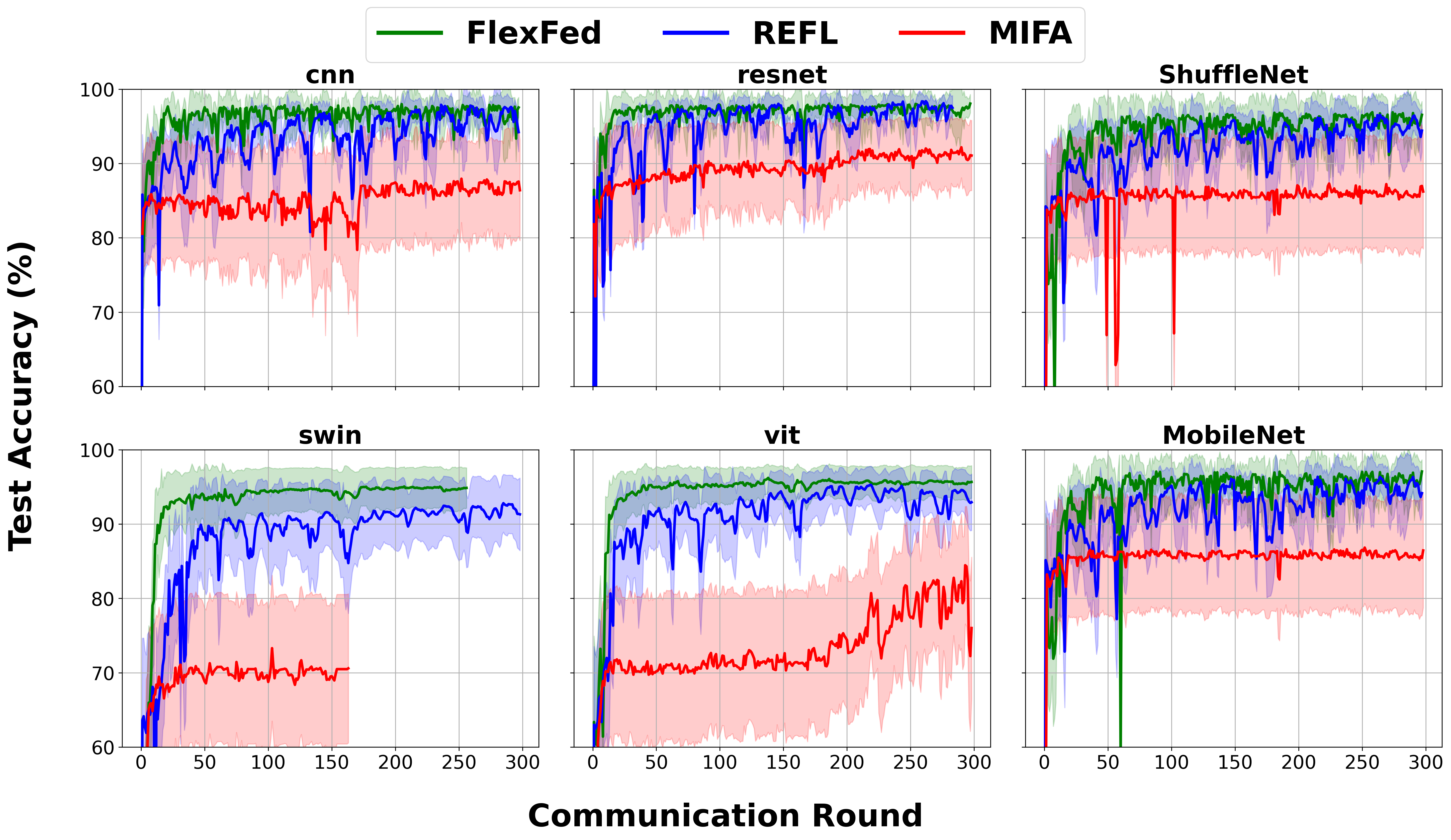}
    \caption{Average Test Accuracy for FlexFed, REFL and  MIFA. The Shaded Regions Represent the Standard Deviation in Performance Across Various HAR Model.}
    \label{fig:FlexFed_REF_MIFA}
\end{figure}

\begin{figure}[H]
    \centering
    \includegraphics[width=1\linewidth, height=4cm]{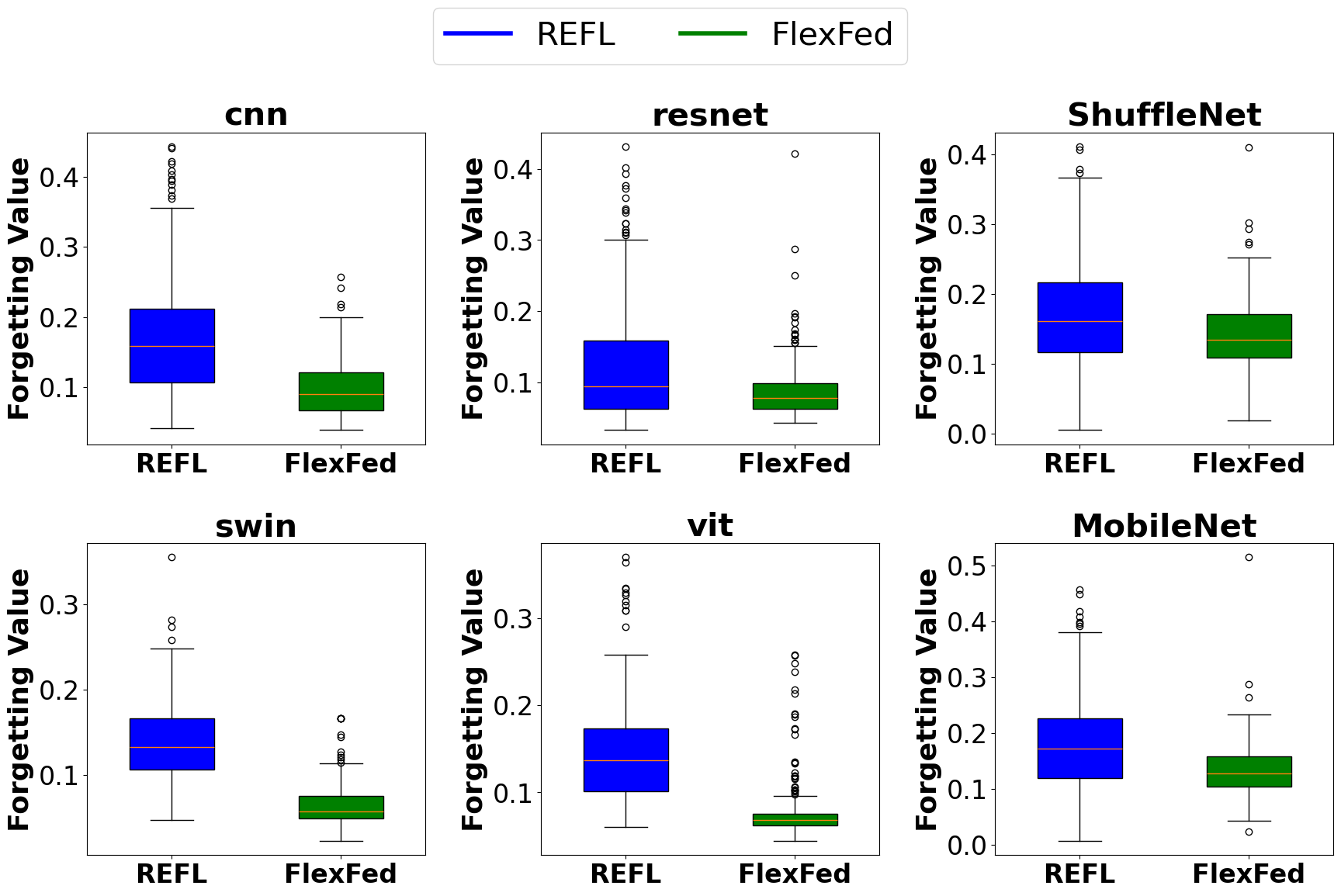}
    \caption{Analysis of Average Forgetting Value per Round Across Various HAR Models.}
    \label{fig:Forgetting_Rate}
\end{figure}

The findings highlight FlexFed's advantages in obtaining faster and more reliable convergence, especially in infrequent labels across various HAR models. Figure~\ref{fig:FlexFed_REF_MIFA} reveals that (i) FlexFed \emph{converges} faster than REFL and MIFA and (ii) constantly achieves the highest accuracy, with stability above 95\% in the majority of rounds for all HAR models. On the other hand, REFL performs competitively, with accuracies of approximately 90-95\%, but with larger fluctuations in accuracy (shaded region represents standard deviation), indicating that REFL is not consistent in its ``not forgetting'' ability. MIFA, while achieving stability, is much less efficient than FlexFed and REFL, with accuracies of about 85\%, as the number of inactive rounds of clients has a big influence on the convergence rate.  We further observe that FlexFed not only delivers higher accuracy but also has a lower variance, showing more consistent performance and less forgetfulness.
\begin{figure}
    \centering
    \includegraphics[width=0.9\linewidth, height=5.5cm]{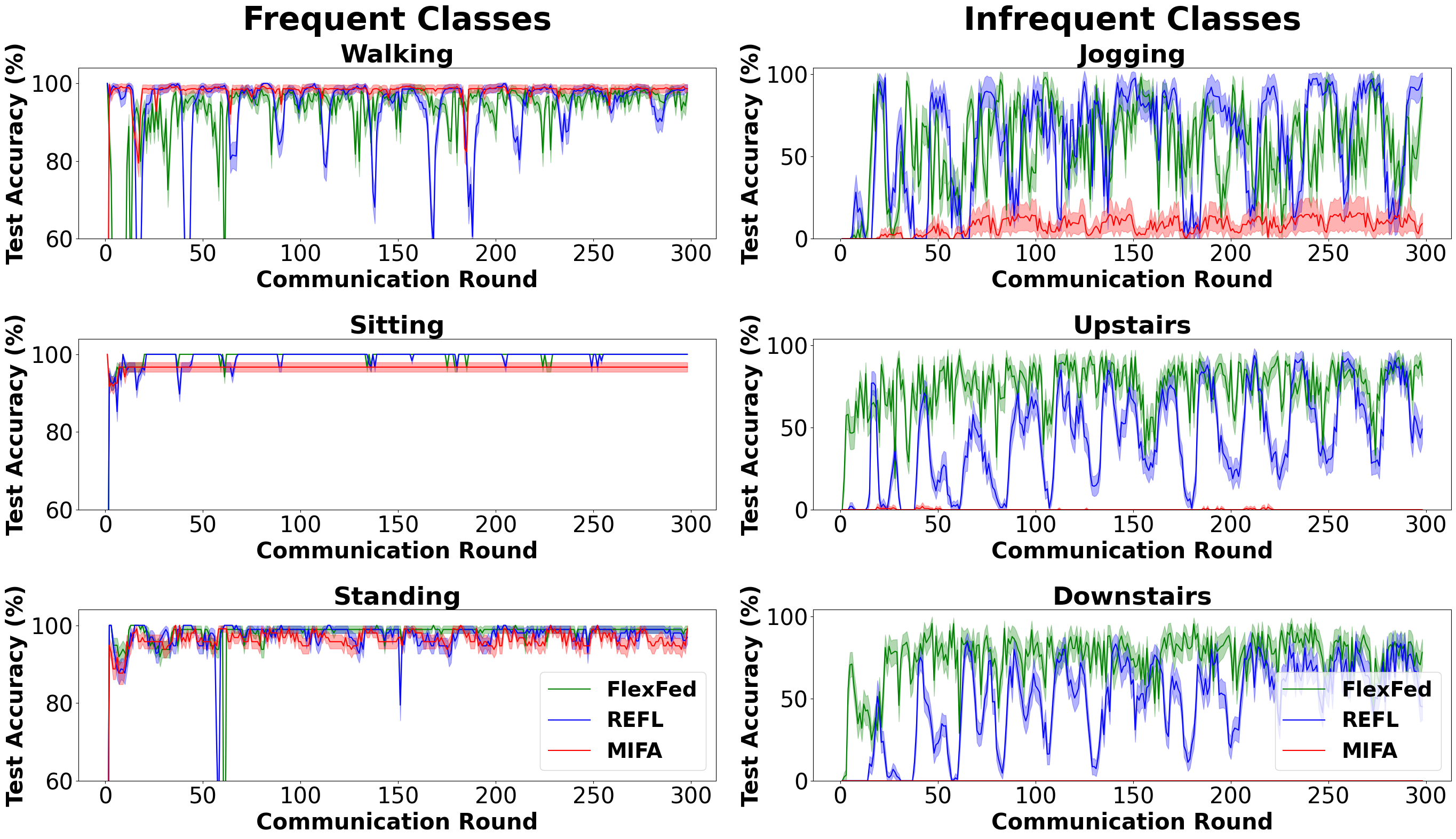}
    \caption{Comparison of the Average Test Accuracy per Activity Class for FlexFed, REFL and  MIFA for HAR CNN Model.}
    \label{fig:Cross-class-Performance}
\end{figure}
Figure~\ref{fig:Cross-class-Performance} compares the accuracies of the three approaches on CNN model, with all performing well on frequent classes, though FlexFed outperforms REFL and MIFA with the lowest variation. Notably, FlexFed achieves convergence in infrequent classes, unlike REFL and MIFA, demonstrating its robustness and effectiveness in mitigating CF for not common labels. Figure~\ref{fig:Forgetting_Rate} depicts the average forgetting values for REFL and FlexFed across all classes and models. FlexFed outperforms REFL by (i) achieving significantly lower forgetting values, indicating more efficient knowledge retention and (ii) exhibiting lower variance, demonstrating more consistent retention of past knowledge compared to REFL.

\begin{figure}
    \centering
    \includegraphics[width=1\linewidth, height=6cm]{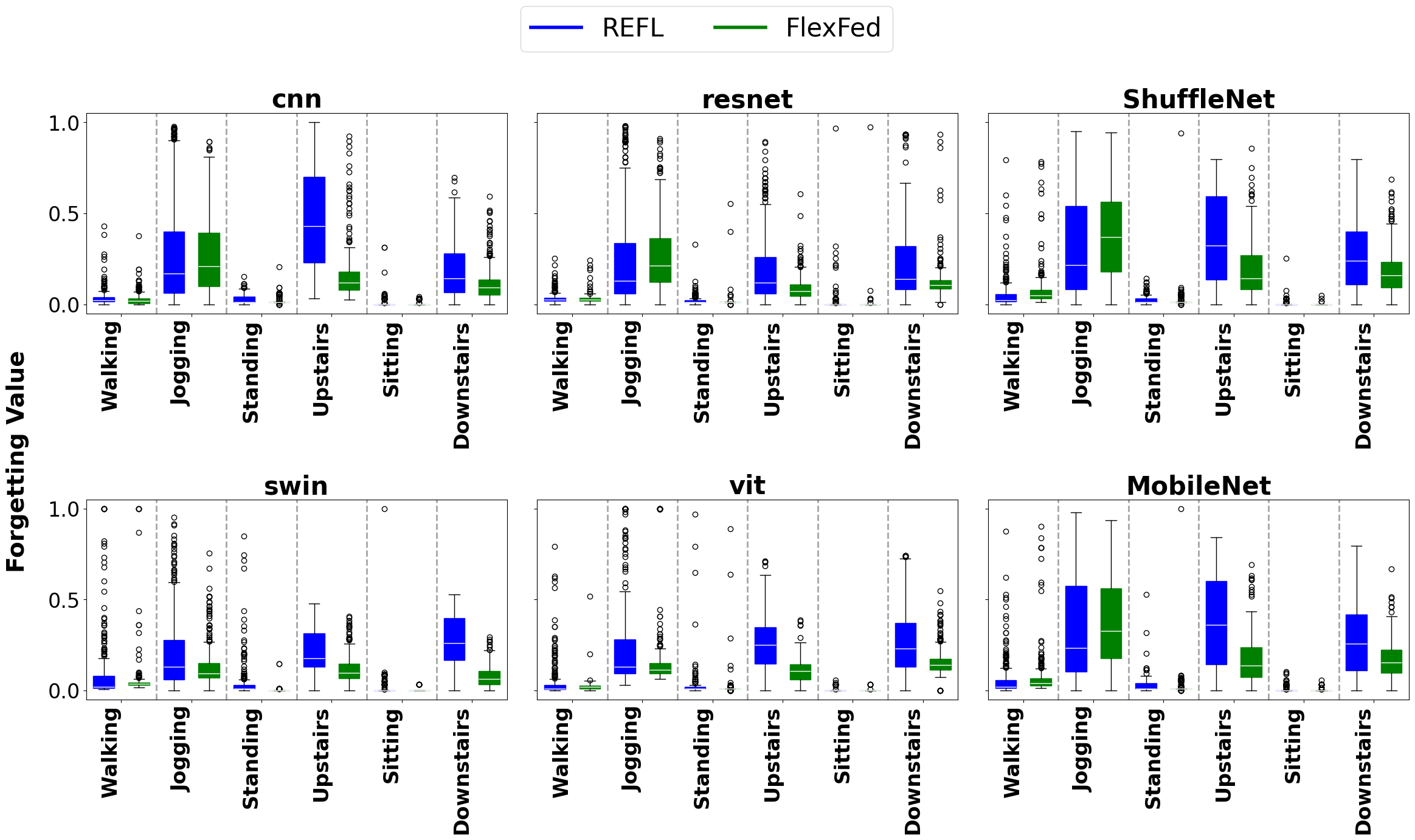}
    \caption{Forgetting Rate per Class Across Various HAR Models.}
    \label{fig:Forgetting_Rate_per_class}
\end{figure}

Figure~\ref{fig:Forgetting_Rate_per_class} shows that, for both frequent and infrequent classes, FlexFed outperforms REFL. The improvement in the forgetting value is even better for infrequent classes, i.e., FlexFed, in general, has a lower average (and distribution of) forgetting values than REFL for every activity, suggesting that FlexFed is better at retaining knowledge. 

\section{Conclusion}
\label{sect:conclusion}
In this work, we investigate the FL problem in pervasive computing environments where (i) data is heterogeneous, (ii) memory of devices is limited and (ii) connectivity to server is intermittent. These issues lead to catastrophic forgetting where models forget previous knowledge when learning new ones. We propose FlexFed, a novel FL approach that handles the forgetting problem in pervasive environments and our results show that FlexFed significantly outperforms two state-of-the-art techniques, namely REFL and MIFA.

\bibliographystyle{plain}
\bibliography{IEEEexample}

\end{document}